\documentclass[runningheads,a4paper]{llncs}

\usepackage{amssymb}
\usepackage{amstext}
\usepackage{graphicx}

\usepackage{url}
\urldef{\mailsa}\path|{alfred.hofmann, ursula.barth, ingrid.haas, frank.holzwarth,|
\urldef{\mailsb}\path|anna.kramer, leonie.kunz, christine.reiss, nicole.sator,|
\urldef{\mailsc}\path|erika.siebert-cole, peter.strasser, lncs}@springer.com|    
\newcommand{\keywords}[1]{\par\addvspace\baselineskip
\noindent\keywordname\enspace\ignorespaces#1}

\begin{document}

\mainmatter  % start of an individual contribution

\title{A distance-based decision in the credal level}

\titlerunning{A distance-based decision in the credal level}

\author{Amira Essaid\inst{1,2} \and Arnaud Martin\inst{2} \and Gr\'egory Smits\inst{2} \and Boutheina Ben Yaghlane\inst{3}}%

\authorrunning{Amira Essaid et al.}

\institute{${}^1$ LARODEC, University of Tunis, ISG Tunis, Tunisia\\
${}^2$IRISA, University of Rennes1, Lannion, France\\
${}^3$ LARODEC, University of Carthage, IHEC Carthage, Tunisia}

\toctitle{Lecture Notes in Computer Science}
\tocauthor{Authors' Instructions}
\maketitle

\begin{abstract}
Belief function theory provides a flexible way to combine information provided by different sources. This combination is usually followed by a decision making which can be handled by a range of decision rules. Some rules help to choose the most likely hypothesis. Others allow that a decision is made on a set of hypotheses. In \cite{essaid14}, we proposed a decision rule based on a distance measure. First, in this paper, we aim to demonstrate that our proposed decision rule is a particular case of the rule proposed in \cite{denoeux97}. Second, we give experiments showing that our rule is able to decide on a set of hypotheses. Some experiments are handled on a set of mass functions generated randomly, others on real databases.
\end{abstract}

\section{Introduction}

Belief function theory \cite{dempster67,shafer76} allows us to represent all kinds of ignorance and offers rules for combining several imperfect information provided by different sources in order to get a more coherent one. The combination process helps to make decisions later. Decision making consists in selecting, for a given problem, the most suitable actions to take. Today, we are often confronted with the challenge of making decisions in cases where information is imprecise or even not available. In \cite{smets94}, Smets proposed the transferable belief model (TBM) as an interpretation of the theory of belief functions. The TBM emphasizes a distinction between knowledge modeling and decision making. Accordingly, we distinguish the credal level and the pignistic level. In the credal level, knowledge is represented as belief functions and then combined. The pignistic level corresponds to decision making, a stage in which belief functions are transformed into probability functions. 

The pignistic probability, the maximum of credibility and the maximum of plausibility are rules that allow a decision on a singleton of the frame of discernment. Sometimes and depending on application domains, it seems to be more convenient to decide on composite hypotheses rather than a simple one. In the literature, there are few works that propose a rule or an approach for making decision on a union of hypotheses \cite{denoeux97,appriou05,martin08}. Recently, we proposed a decision rule based on a distance measure \cite{essaid14}. This rule calculates the distance between a combined mass function and a categorical one. The most likely hypothesis to choose is the hypothesis whose categorical mass function is the nearest to the combined one. 
%We extend the rule with no categorical mass functions. 

The main topic of this paper is to demonstrate that our proposed decision rule is a particular case of that detailed in \cite{denoeux97} and to extend our rule so that it becomes able to give decisions even with no categorical mass functions. We present also our experiments on mass functions generated randomly as well as on real databases.

The remainder of this paper is organized as follows: in section 2 we recall the basic concepts of belief function theory. Section 3 presents our decision rule based on a distance measure proposed in \cite{essaid14}. In section 4, we demonstrate that our proposed rule is a particular case of that proposed in \cite{denoeux97}. Section 5 presents experiments and the main results. Section 6 concludes the paper.

\section{The theory of belief functions}

The theory of belief functions \cite{dempster67,shafer76} is a general mathematical framework for representing beliefs and reasoning under uncertainty. In this section, we recall some concepts of this theory.

The \textit{frame of discernment} $\Theta = \left\{\theta_1, \theta_2, \ldots, \theta_n\right\}$ is a set of \textit{n} elementary hypotheses related to a given problem. These hypotheses are exhaustive and mutually exclusive.
The power set of $\Theta$, denoted by $2^\Theta$ is the set containing singleton hypotheses of $\Theta$, all the disjunctions of these hypotheses as well as the empty set.

The Basic belief assignment \textit{(bba)}, denoted by \textit{m} is a mass function defined on $2^\Theta$. It affects a value from $\left[0,1\right]$ to each subset. It is defined as:
\begin{equation}
\sum_{A \subseteq 2^\Theta} m(A)=1.
\end{equation}
A focal element \textit{A} is an element of $2^\Theta$ such that $m(A)>0$. A categorical bba is a bba with a unique focal element such that $m(A) = 1$. When this focal element is a disjunction of hypotheses then the bba models imprecision.

Based on the basic belief assignment, other belief functions (credibility function ad plausibility function) can be deduced.
\begin{itemize}
	\item Credibility function \textit{bel(A)} expresses the total belief that one allocates to $A$. It is a mapping from elements of $2^\Theta$ to $\left[0,1\right]$ such that:
			\begin{equation}
				\begin{tabular}{l}
					$bel(A) = \displaystyle{\sum_{B\subseteq A, B\neq\emptyset}}m(B)$.\\
				\end{tabular}
			\end{equation}

	\item Plausibility function \textit{pl(A)} is defined as:
				\begin{equation}
					\begin{tabular}{l}
					$pl(A) = \displaystyle{\sum_{A\cap B\neq\emptyset}}m(B)$.\\
					\end{tabular}
			\end{equation}
			The plausibility function measures the maximum amount of belief that supports the proposition $A$ by taking into account all the elements that do not contradict. The value \textit{pl(A)} quantifies the maximum amount of belief that might support a subset $A$ of $\Theta$.
\end{itemize}

%A variety of combination rules exist \cite{dempster67,yager87,dubois88,smets93}. The first combination rule has been proposed by \cite{dempster67} and is defined as:
%
%\begin{equation}%Equation(1.26)
%m_{1\oplus 2}(A)=
%\left\{
%\begin{tabular}{ll}
%$\frac{\displaystyle \sum _{B\cap C=A} m_1(B)\times m_2(C)}{1-\displaystyle \sum _{B\cap C=\emptyset} m_1(B) \times m _2(C)}$ & $\forall A\subseteq\Theta,\hspace{0.1cm}A\neq \emptyset$\\
%0&if $A=\emptyset$\\
%\end{tabular}
%\right.
%\end{equation}
%

The theory of belief function is a useful tool for data fusion. In fact, for a given problem and for the same frame of discernment, it is possible to get a mass function synthesizing knowledge from separate and independent sources of information through applying a combination rule. Mainly, there exists three modes of combination:
\begin{itemize}
	\item Conjunctive combination is used when two sources are distinct and fully reliable. In \cite{smets90}, the author proposed the conjunctive combination rule which is defined as:
	\begin{equation}
		\label{SmetsRule}
			m_{1\textcircled{\scriptsize{$\cap$}}2}(A)=\sum _{B\cap C=A} m_1(B)\times m_2(C).
	\end{equation}

The Dempster's rule of combination \cite{dempster67} is a normalized form of the rule described previously and is defined as:
	\begin{equation}%Equation(1.26)
\label{Equation(4)}
m_{1\oplus 2}(A)=
\left\{
\begin{tabular}{ll}
$\frac{\displaystyle{\sum _{B\cap C=A}} m_1(B)\times m_2(C)}{1-\displaystyle{\sum _{B\cap C=\emptyset}} m_1(B)\times 
m _2(C)}$&$\forall A\subseteq\Theta,\hspace{0.1cm}A\neq \emptyset$\\
0&$\text{if}\hspace{0.1cm}A=\emptyset$\\
\end{tabular}
\right.
\end{equation}
	This rule is normalized through $1-\displaystyle{\sum _{B\cap C=\emptyset}} m_1(B)\times m _2(C)$ and it works under the closed world assumption where all the possible hypotheses of the studied problem are supposed to be enumerated on $\Theta$.
	
	\item Disjunctive combination: In \cite{smets93}, Smets introduced the disjunctive combination rule which combines mass functions when an unknown source is unreliable. This rule is defined as:
		\begin{equation}
			\label{Equation(5)}
				m_{1\textcircled{\scriptsize{$\cup$}}2}(A)=\sum _{B\cup C=A} m_1(B)\times m_2(C)
		\end{equation}

	\item Mixed combination: In \cite{dubois88}, the authors proposed a compromise in order to consider the benefits of the two combination modes previously described. This combination is given for every $A\in 2^\Theta$ by the following formula:

\begin{equation}
\label{eq19:mixedCombination}
\left\{
\begin{tabular}{ll}
$\displaystyle {m_{DP}(A)=m_{1\textcircled{\scriptsize{$\cap$}}}(A) + \sum_{B \cap C = \emptyset, B \cup C = A}{m_1 (B) m_2(C)}}$&$\forall A\in 2^\Theta,\hspace{0.1cm} A\neq\emptyset$\\

$m_{DP}(\emptyset)=0$\\
\end{tabular}
\right.
\end{equation}

\end{itemize}

\section{Decision Making in the theory of belief functions}
%Belief combination helps to make decision which consists in selecting, for a given problem, the most suitable action to handle.
In the transferable belief model, decision is made on the pignistic level where the  belief functions are transformed into a probability function, named \textit{pignistic probability}. This latter, noted as $BetP$ is defined for each X $\in 2^\Theta$, $X \neq 0$ as:
\begin{equation}%Equation(1.26)
	\begin{tabular}{ll}
		$\displaystyle{betP(X) = \sum_{Y \in 2^\Theta, Y\neq\emptyset}\frac{|X \cap Y|}{|Y|}\frac{m(Y)}{1 - m(\emptyset)}}$
	\end{tabular}
\end{equation}

where $|Y|$ represents the cardinality of $Y$.

Based on the obtained pignistic probability, we select the most suitable hypothesis with the maximum \textit{BetP}. This decision results from applying tools of decision theory \cite{denoeux97}. In fact, if we consider an entity represented by a feature vector \textit{x}. \textit{A} is a finite set of possible actions $A =\left\{a_1, \ldots, a_N\right\}$ and $\Theta$ a finite set of hypotheses, $\Theta =\left\{\theta_1, \ldots, \theta_M\right\}$. An action $a_j$ corresponds to the action of choosing the hypothesis $\theta_j$. But, if we select $a_i$ as an action whereas the hypothesis to be considered is rather $\theta_j$ then the loss occurred is $\lambda(a_i | \theta_j)$. The expected loss associated with the choice of the action $a_i$ is defined as:

\begin{equation}%Equation(1.26)
	\begin{tabular}{ll}

		$\displaystyle{R_{betP}(a_i| x) = \sum_{\theta_j\in\Theta}\lambda(a_i| \theta_j)BetP(\theta_j)}$.
	\end{tabular}
\end{equation}
Then, the decision consists in selecting the action which minimizes the expected loss. In addition to minimizing pignistic expected loss, other risks are presented in \cite{denoeux97}.

Decision can be made on composite hypotheses \cite{appriou05,martin08}. We present in this paper the Appriou's rule \cite{appriou05} which helps to choose a solution of a given problem by considering all the elements contained in $2^\Theta$. This approach weights the decision functions (maximum of credibility, maximum of plausibility and maximum of pignistic probability) by an utility function depending on the cardinality of the elements. $A \in 2^\Theta$ is chosen if: 

\begin{equation}
			\displaystyle{A= \mathop{argmax} _{X\in 2^\Theta} (m_d(X)pl(X))}
\end{equation}
where $m_d$ is a mass defined by: 
	\begin{equation}
			m_d(X) = K_d\lambda_{X}\left(\frac{1}{|X|^r}\right)
		\end{equation}
The value \textit{r} is a parameter in $\left[0, 1\right]$ helping to choose a decision which varies from a total indecision when \textit{r} is equal to 0 and a decision based on a singleton when \textit{r} is equal 1. $\lambda_{X}$ helps to integrate the lack of knowledge about one of the elements of \textit{$2^\Theta$}. $K_d$ is a normalization factor and $pl(X)$ is a plausibility function.

In the following, we present our decision rule based on a distance measure.

\section{Decision rule based on a distance measure}

In \cite{essaid14}, we proposed a decision rule based on a distance measure. It is defined as:

\begin{equation}
	A = \mathop{argmin}(d(m_{comb}, m_A))
\end{equation}

This rule aims at deciding on a union of singletons. It is based on the use of categorical bba which helps to adjust the degree of imprecision that has to be kept when deciding. Depending on cases, we can decide on unions of two elements or three elements, etc. The rule calculates the distance between a combined bba $m_{comb}$ and a categorical one  $m_A$ . The minimum distance is kept and the decision corresponds to the categorical bba's element having the lowest distance with the combined bba. The rule is applied as follows:
\begin{itemize}
	\item We consider the elements of $2^\Theta$. In some applications, $2^\Theta$ can be of a large cardinality. For this reason, we may choose some elements to work on. For example, we can keep the elements of $2^\Theta$ whose cardinality is less or equal to 2.
	\item For each selected element, we construct its corresponding categorical bba.
	\item Finally, we apply Jousselme distance \cite{jousselme01} to calculate the distance between the combined bba and a categorical bba. The distance with the minimum value is kept. The most likely hypothesis to select is the hypothesis whose categorical bba is the nearest to the combined bba.
\end{itemize}

Jousselme distance is defined for two bbas $m_1$ and $m_2$ as follows:
\begin{equation}
d(m_1,m_2)=\sqrt{\frac{1}{2}(m_1-m_2)^t\underline{\underline{D}}(m_1-m_2)}
\end{equation}
where \underline{\underline{D}} is a matrix based on Jaccard distance as a similarity measure between focal elements. This matrix is defined as:
\begin{equation}
D(A,B)=\left\{
\begin{tabular}{ll}
1&if A=B=$\emptyset$\\
$\frac{\mid A\cap B\mid}{\mid A\cup B\mid}$&$\forall A,B \in 2^\Theta$\\
\end{tabular}
\right.
\end{equation}

In this paper, we propose to apply the rule through two different manners:
\begin{itemize}
	\item \textit{Distance type 1} is calculated with categorical bbas ($m(A) =1$) for all elements of $2^\Theta$ except $\Theta$ to have an imprecise result rather than a total ignorance.
	\item  \textit{Distance type 2} is calculated with simple bbas such as $m(A)=\alpha$, $m(\Theta)=1-\alpha$.
\end{itemize}
In the following, we show that our proposed rule can be seen as a particular case of that proposed in section 3. 

Jousselme distance can be written as:
\begin{equation}
 d(m_1,m_2)= \frac{1}{2} \sum_{Y \subseteq \Theta}\sum_{X\subseteq \Theta} \frac{|X\cap Y|} {|X \cup Y|}m(X)m(Y)
\end{equation}

If we consider the expected loss of choosing $a_i$, then it can be written as:
\begin{equation}%Equation(1.26)
	\begin{tabular}{ll}

		$\displaystyle{R_{betP}(a_i| x) = \sum_{Y \in \Theta}\lambda(a_i|Y)BetP(Y)}$.\\
		
		$\displaystyle{R_{betP}(a_i| x) = \sum_{Y \in \Theta}\lambda(a_i|Y) \sum_{X\in\Theta}\frac{|X \cap Y|}{|X|}\frac{m(X)}{1 - m(\emptyset)}}$.\\
		$\displaystyle{R_{betP}(a_i| x) = \sum_{Y \in \Theta}\sum_{X\in\Theta}\lambda(a_i|Y) \frac{|X \cap Y|}{|X|}\frac{m(X)}{1 - m(\emptyset)}}$.\\
		
	\end{tabular}
\end{equation}

The equation relative to decision is equal to that for the risk for a value of $\lambda$ that has to be equal to:
\begin{equation}
\begin{tabular}{ll}
	$\displaystyle{\lambda(a_i|Y)= \frac{|X|(1-m(\emptyset))}{|X\cup Y|} m(X)}$
\end{tabular}
\end{equation}

In this section, we showed that for a particular value of $\lambda$, our proposed decision rule can be considered as a particular case of that proposed in \cite{denoeux97}. In the following section, we give experiments and present comparisons between our decision rule based on a distance measure and that presented in \cite{appriou05}.

\section {Experiments}

\subsection{Experiments on generated mass functions}

We tested the proposed rule \cite{essaid14} on a set of mass functions generated randomly. To generate the bbas, one needs to specify the cardinality of the frame of discernment, the number of mass functions to be generated as well as the number of focal elements. The generated bbas are then combined. We use the Dempster's rule of combination, the disjunctive rule and the mixed rule. Suppose we have a frame of discernment represented as $\Theta = \left\{\theta_1, \theta_2, \theta_3 \right\}$ and three different sources for which we generate their corresponding bbas as given in Table~\ref{tab:bbas}.

\begin{table}
\caption{Three sources with their bbas}
\label{tab:bbas}  
%\begin{tabular}{ccccccc}
\begin{center}
\begin{tabular}{|c|c|c|c|}
\hline 
& $S_1$ & $S_2$ & $S_3$\\\hline
$\theta_1$ & 0.410 & 0.223 & 0.034 \\
$\theta_2$ & 0.006 & 0.108 & 0.300 \\
$\theta_1\cup\theta_2$ & 0.039 &  0.027 &0.057 \\
$\theta_3$ & 0.026 & 0.093 & 0.128 \\
$\theta_1\cup\theta_3$ & 0.094 & 0.062 & 0.04\\
$\theta_2\cup\theta_3$ & 0.199 & 0.153 & 0.004\\ 
$\theta_1\cup\theta_2\cup\theta_3$  & 0.226&0.334&0.437\\

\hline 
\end{tabular}

\end{center}
\end{table}

We apply combination rules and we get the results illustrated in Table~\ref{tab:comb}.
\begin{table}
\caption{Combination results}
\label{tab:comb}  
%\begin{tabular}{ccccccc}
\begin{center}
\begin{tabular}{|c|c|c|c|}
\hline 
& Dempster rule & Disjunctive rule & Mixed rule\\\hline
$\theta_1$ & 0.369 & 0.003 & 0.208 \\
$\theta_2$ & 0.227 & 0 & 0.128 \\
$\theta_1\cup\theta_2$ & 0.025&  0.061 &0.075 \\
$\theta_3$ & 0.168  & 0 & 0.094 \\
$\theta_1\cup\theta_3$ & 0.049  & 0.037 & 0.064\\
$\theta_2\cup\theta_3$ & 0.103  & 0.035 & 0.093\\ 
$\theta_1\cup\theta_2\cup\theta_3$  & 0.059 &0.864&0.338\\
\hline 
\end{tabular}
\end{center}
\end{table}

\begin{table}
\caption{Decision results}
\label{decision}
\begin{center}
\begin{tabular}{|c|c|c|c|}
\hline
 & Pignistic  & Appriou rule & Rule based on\\
 &  Probability &  & distance measure\\
\hline
 Dempster rule      & $\theta_1$ & $\theta_1 \cup \theta_2$ & $\theta_1 $\\
Disjunctive rule   & $\theta_1$ & $\theta_1$ & $\theta_1 \cup \theta_2$  \\
Mixed rule         & $\theta_1$ & $\theta_1$ & $\theta_1 \cup \theta_2$   \\

\hline

\end{tabular}
\end{center}
\end{table}

Once the combination is performed, we can make decision. In Table~\ref{decision}, we compare between the results of three decision rules, namely the pignistic probability, the Appriou's rule with \textit{r} equal to 0.5 as well as our proposed decision rule based on distance measure.

Table~\ref{decision} shows the decision results obtained after applying some combination rules. We depict from this table that not all the time the rule proposed by Appriou gives a decision on a composite hypotheses. In fact, as shown in Table~\ref{decision}, the application of disjunctive rule as well as the mixed rule lead to a decision on a singleton which is $\theta_1$. This is completely different from what we obtain when we apply our proposed rule which promotes a decision on union of singletons when combining bbas. The obtained results seems to be convenient especially that the disjunctive and the mixed rules help to get results on unions of singletons.

\subsection{Experiments on real databases}
To test our proposed decision rule, we do some experiments on real databases (IRIS\footnote{http://archive.ics.uci.edu/ml/datasets/Iris} and HaberMan's survival\footnote {http://archive.ics.uci.edu/ml/datasets/Haberman$\%$27s+Survival}). Iris is a dataset contaning 150 instances, 4 attributes and 3 classes where each class refers to a type of iris plant. HaberMan is a dataset containing results study conducted at the University of Chicago's Billings Hospital on the survival of patients who had undergone surgery for breast cancer. This dataset contains 306 instances, 3 attributes and 2 classes (1: patient survived 5 years or longer, 2: patient died within 5 years). For the classification, our experiments are handled in two different manners. 
\begin{itemize}
	\item First, we apply the $k$-NN classifier \cite{denoeux95}. The results are illustrated in a confusion matrix as shown in Table~\ref{tab:Iris} (left side).
	\item Second, we modify the $k$-NN classifier's algorithm based on the use of Dempster rule of combination, to make it able to combine belief functions through the mixed rule. Then, Appriou's rule and our proposed decision rule are applied to make decision.
	Results are illustrated in Table ~\ref{tab:Iris}.
\end{itemize}

\begin{table}
\caption{Confusion Matrices for Iris}
\label{tab:Iris}  
\begin{tabular}{ccccccc}
\begin{tabular}{|c|c|c|c|}
\hline
\multicolumn{4}{|c|}{$k$-NN classifier}\\
\hline 
& $\theta_1$ & $\theta_2$ &$\theta_3$ \\
\hline
$\theta_1$        & 11 & 0  & 0 \\
$\theta_2$   			& 0  & 11 & 2  \\
$\theta_3$        & 0  & 0  & 16  \\
\hline
\end{tabular}
\quad

\begin{tabular}{|c|c|c|c|c|c|c|}
\hline
\multicolumn{7}{|c|}{Appriou's rule}\\
\hline
  & $\theta_1$ & $\theta_2$&$\theta_1\cup\theta_2$&$\theta_3$&$\theta_1\cup\theta_3$&$\theta_2\cup\theta_3$ \\
\hline
 $\theta_1$       & 11 & 0  & 0 & 0  & 0 & 0  \\
$\theta_2$   	    & 0  & 15 & 0 & 0  & 0 & 0 \\
$\theta_3$        & 0  & 1  & 0 & 13 & 0 & 0  \\

\hline
\end{tabular}
\quad

\begin{tabular}{|c|c|c|c|c|c|c|}
\hline 
\multicolumn{7}{|c|}{Our decision rule}\\
\hline
 & $\theta_1$ & $\theta_2$&$\theta_1\cup\theta_2$&$\theta_3$&$\theta_1\cup\theta_3$&$\theta_2\cup\theta_3$ \\
\hline
$\theta_1$        & 10 & 0  & 0 & 0   & 0 & 0  \\
$\theta_2$   	    & 0  & 12 & 0 & 2   & 0 & 1  \\
$\theta_3$        & 0  & 0  & 0 & 13  & 0 & 2  \\
\hline
\end{tabular}
%\end{table}
\end{tabular}
\end{table}

The same tests are done for HaberMan's survival dataset. The results of applying $k$-NN classifier, Appriou's rule and our decision rule are given respectively in Table~\ref{tab:HaberMan}. For the classification of 40 sets chosen randomly from Iris, we remark that with the $k$-NN classifier, all the sets having $\theta_1$ and $\theta_3$ as corresponding classes are well classified and only two originally belonging to class $\theta_2$ were classified as $\theta_3$. Appriou's rule gives a good classification for sets originally belonging to classes $\theta_1$ and $\theta_2$ and thus promoting a result on singletons rather than on a union of singletons.

Considering the results obtained when applying our decision rule based on a distance type 1, we note that only 2 sets are not well classified and that 3 have $\theta_2\cup\theta_3$ as a class. The obtained results are good because our method is based on an imprecise decision which is underlined by the fact of obtaining $\theta_2\cup\theta_3$ as a class.  

\begin{table}
\caption{Confusion Matrices for HaberMan's survival}
\label{tab:HaberMan}  
\begin{tabular}{ccccccc}
\begin{tabular}{|c|c|c|}
\hline 
 \multicolumn{3}{|c|}{$k$-NN classifier}\\
\hline
& $\theta_1$ & $\theta_2$  \\
\hline
$\theta_1$        & 34  & 4  \\
$\theta_2$   			& 12  & 6 \\\hline
\end{tabular}
\quad \quad

\begin{tabular}{|c|c|c|c|}
\hline
\multicolumn{4}{|c|}{Appriou's rule}\\
\hline
 & $\theta_1$ & $\theta_2$& $\Theta$ \\
\hline
$\theta_1$      & 34 & 4  & 0  \\
$\theta_2$   	  & 12 & 6  & 0 \\
\hline

\end{tabular}
\quad \quad

\begin{tabular}{|c|c|c|c|}
\hline
\multicolumn{4}{|c|}{Our decision rule}\\
\hline
 & $\theta_1$ & $\theta_2$& $\Theta$ \\
\hline
$\theta_1$     & 34 & 4  &  0 \\
$\theta_2$   	 & 12 	& 6  &  0 \\
\hline

\end{tabular}
\end{tabular}
\end{table}

Considering HaberMan's survival dataset, we note that the $k$-NN classifier, Appriou's rule as well as our decision rule give the same results where among the sets originally belonging to $\theta_1$, 34 are  well classified and among the 18 belonging to $\theta_2$, only 6 are well classified. We obtain the same results as the other rules because the HaberMan's survival dataset has only two classes and our method is based on getting imprecise decisions and excluding the ignorance.

All the experiments given previously are based on the use of distance type 1. The results shown below are based on distance type 2. In fact, we consider a simple bba and each time, we assign a value $\alpha$ to an element of $2^\Theta$. The tested rule on Iris as illustrated in Table~\ref{tab:Rates} (left side) gives better results with an $\alpha<0.8$. In addition to that, we obtained decisions on a union of singletons. The tests done on HaberMan's survival as given in Table~\ref{tab:Rates} (right side) shows that with $\alpha>0.5$, we obtain a better rate of good classification although we did not obtain a good classification for the class $\theta_2$ and no set belongs to $\Theta$. We aim in the future to make experiments on other datasets because HaberMan's survival, for example, does only have 2 classes, so we do not have enough imprecise elements.

\begin{table}
\caption{Rates of good classification}
\label{tab:Rates}  
\begin{tabular}{cc}
\begin{tabular}{|c|c|c|}
\hline 
 & $\alpha<0.8$ & $\alpha>=0.8$\\
\hline
Iris  & 0.95  & 0.675  \\
\hline
\end{tabular}
\quad \quad
  
\begin{tabular}{|c|c|c|c|}
\hline
 & $\alpha<=0.2$ & $\alpha \in \left[0.3,0.5\right]$ & $\alpha>0.5$\\
\hline
HaberMan's survival & 0.786  & 0.803 & 0.821  \\
\hline
\end{tabular}
\end{tabular}
\end{table}

\section{Conclusion}
In this paper, we presented a rule based on a distance measure. This decision rule helps to choose the most likely hypothesis based on the calculation of the distance between a combined bba and a categorical bba. The aim of the proposed decision rule is to give results on composite hypotheses. In this paper, we demonstrated that our proposed rule can be seen as a particular case of that proposed in \cite{denoeux97}. We presented also the different experiments handled on generated mass functions as well as on real databases.

\end{document}